\def\year{2021}\relax
\documentclass[letterpaper]{article} 
\usepackage{aaai21}  
\usepackage{times}  
\usepackage{helvet} 
\usepackage{courier}  
\usepackage[hyphens]{url}  
\usepackage{graphicx} 
\usepackage{natbib}  
\usepackage{caption} 
\usepackage{graphicx}
\usepackage{xspace}
\usepackage{paralist}
\usepackage{booktabs}
\usepackage{multirow}
\usepackage{amsmath}
\usepackage{amsfonts}
\usepackage{pifont}
\usepackage{amssymb}
\usepackage{mathrsfs}
\usepackage{algorithm,algpseudocode}
\usepackage{mdwlist}

\newcommand{\cmark}{\ding{51}}
\newcommand{\xmark}{\ding{55}}
\usepackage{enumitem}
\usepackage{color}
\usepackage{array}
\definecolor{darkblue}{rgb}{0,0,0.5}
\usepackage[switch]{lineno}

\newcommand{\method}{COSTT\xspace}

\newcommand{\xx}{\mathbf{x}}
\newcommand{\uu}{\mathbf{u}}
\newcommand{\zz}{\mathbf{z}}
\newcommand{\yy}{\mathbf{y}}
\newcommand{\hh}{\mathbf{h}}

\title{Consecutive Decoding for Speech-to-text Translation}

\author {
    Qianqian Dong, \textsuperscript{\rm 1,2}\thanks{The work was done while QD was a research intern at ByteDance AI Lab.}
    Mingxuan Wang, \textsuperscript{\rm 3}
    Hao Zhou, \textsuperscript{\rm 3}
    Shuang Xu, \textsuperscript{\rm 1}
    Bo Xu, \textsuperscript{\rm 1,2}
    Lei Li \textsuperscript{\rm 3}\\
}
\affiliations {
    \textsuperscript{\rm 1} Institute of Automation, Chinese Academy of Sciences (CASIA), Beijing, 100190, China \\
    \textsuperscript{\rm 2} School of Artificial Intelligence, University of Chinese Academy of Sciences, Beijing, 100049, China \\
    \textsuperscript{\rm 3} ByteDance AI Lab, China \\
    \{dongqianqian2016, shuang.xu, xubo\}@ia.ac.cn, 
    \{wangmingxuan.89, zhouhao.nlp, lileilab\}@bytedance.com
}

\begin{document}

\maketitle

\begin{abstract}
Speech-to-text translation (ST), which directly translates the source language speech to the target language text, has attracted intensive attention recently. However, the combination of speech recognition and machine translation in a single model poses a heavy burden on the direct cross-modal cross-lingual mapping. To reduce the learning difficulty, 
we propose COnSecutive Transcription and Translation (\method), an integral approach for speech-to-text translation. 
The key idea is to generate source transcript and target translation text with a single decoder. 
It benefits the model training so that additional large parallel text corpus can be fully exploited to enhance the speech translation training. 
Our method is verified on three mainstream datasets, including Augmented LibriSpeech English-French dataset, IWSLT2018 English-German dataset, and TED English-Chinese dataset. 
Experiments show that our proposed \method outperforms or on par with the previous state-of-the-art methods on the three datasets. 
We have released our code at \url{https://github.com/dqqcasia/st}.

\end{abstract}

\section{Introduction}
\label{sec:intro}

Speech translation (ST) aims at translating from source language speech into the target language text. 
Traditionally, it is realized by cascading an automatic speech recognition (ASR) and a machine translation (MT) ~\citep{sperber2017neural,sperber2019self,zhang2019lattice,beck2019neural,cheng2019breaking}.
Recently, end-to-end ST has attracted much attention due to its appealing properties, such as lower latency, smaller model size, and less error accumulation ~\citep{liu2019end,liu2018ustc,weiss2017sequence,berard2018end,duong2016attentional,jia2019leveraging}.

Although end-to-end systems are very promising, cascaded systems still dominate practical deployment in industry. The possible reasons are:
\begin{inparaenum}[\it a)]
    \item Most research work compared cascaded
    and end-to-end models under identical data situations. However, in practice, the cascaded system can benefit from the accumulating independent speech recognition or machine translation data, while the end-to-end system still suffers from the lack of end-to-end corpora. 
    \item  Despite the advantage of reducing error accumulation, 
    the end-to-end system has to integrate multiple complex deep learning tasks into a single model to solve the task, which introduces heavy burden for the cross-modal and cross-lingual mapping. 
    Therefore, it is still an open problem whether end-to-end models or cascaded models are generally stronger.
\end{inparaenum}

We argue that a desirable ST model should take advantages of both end-to-end and cascaded models and acquire the practically acceptable capabilities as follows: 
\begin{inparaenum}[\it a)]
    \item it should be end-to-end to avoid error accumulation;
    \item it should be flexible enough to leverage large-scale independent ASR or MT data.
\end{inparaenum}
At present, few existing end-to-end models can meet all these goals. 
Most studies resort to pre-training or multitask learning to bridge the benefits of cascaded and end-to-end models  \citep{bansal2018pre,sung2019towards,sperber2019attention}. 
A de-facto framework usually initializes the ST model with the encoder trained from ASR data (i.e. source audio and source text pairs) and then fine-tunes on a speech translation dataset to make the cross-lingual translation.
However, it is still challenging for these methods to leverage the bilingual MT data, due to the lack of intermediate text translating stage.  

Our idea is motivated by two motivating insights from ASR and MT models. 
\begin{inparaenum}[\it a)]
    \item A branch of ASR models has intermediate steps to extract acoustic feature and decode phonemes, before emitting transcription; 
    and 
    \item Speech translation can benefit from decoding the source speech transcription in addition to the target translation text. 
\end{inparaenum}
We propose \method, a unified speech translation framework with consecutive decoding for jointly modeling speech recognition and translation.
\method consists of two phases, an acoustic-semantic modeling phase (AS) and a transcription-translation modeling phase (TT). 
The AS phase accepts the speech features and generates compressed acoustic representations.  
For TT phases, we jointly model both the source and target text in a single shared  decoder, which directly generates the speech text sequence and the translation sequence at one pass. 
This architecture is closer to cascaded translation while maintaining the benefits of end-to-end models.
The combination of the AS and the first-part output of the TT phase serves as an ASR model; the TT phase alone serves as an MT model; while the whole makes an end-to-end speech translation by ignoring the first-part of TT output.
Simple and effective, \method is powerful enough to cover the advantage of ASR, MT, and ST models simultaneously.

The contributions of this paper are as follows:
\begin{inparaenum}[\it 1)]
\item We propose \method, a unified training framework with consecutive decoding which bridges the benefits of both cascaded and end-to-end models.
\item As a benefit of explicit multi-phase modeling, \method facilitates the use of parallel bilingual text corpus, which is difficult for traditional end-to-end ST models.
\item \method can achieve state-of-the-art results on three popular benchmark datasets with less external data.
\end{inparaenum}

\section{Related Work}
\label{sec:related}
\begin{figure*}[htb]
    \centering
    \includegraphics[width=1.0\textwidth]{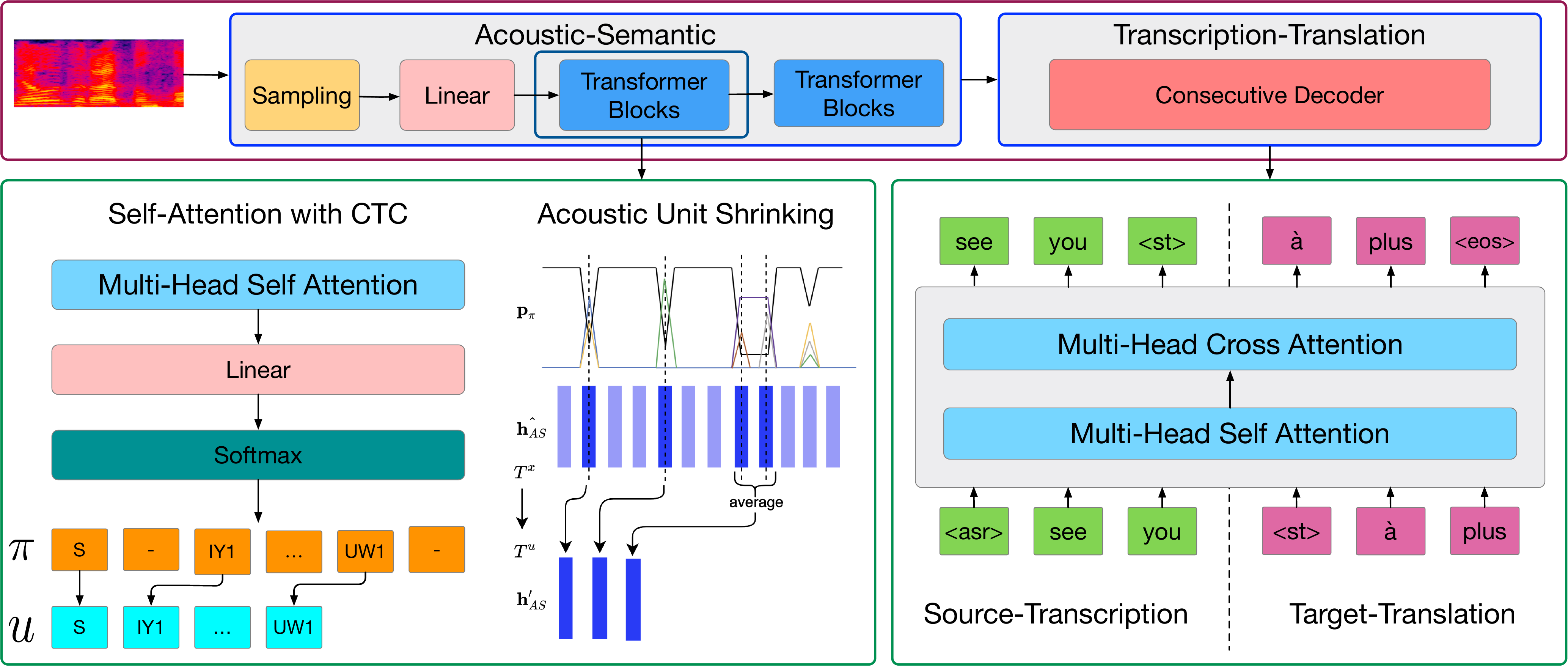}
    \caption{Overview of the proposed \method. It consists of two phases, an acoustic-semantic modeling phase (AS) and a transcription-translation phase (TT). 
    During AS phase, CTC loss is adopted supervised by phoneme labels corresponding to source-text. 
    The TT phase decodes source-text and target-text in a single sequence consecutively.}
    \label{fig:overview}
\end{figure*}

For speech translation, there are two main research paradigms, the cascaded system and the end-to-end model~\citep{jan2018iwslt,jan2019iwslt}.

For cascaded system, the most concerned point is how to avoid early decisions, relieve error propagation and better integrate the separately trained ASR and MT modules. To relieve the problem of error propagation and tighter couple cascaded systems:
\begin{inparaenum}[\it a)]
\item Robust translation models~\citep{cheng2018towards,cheng2019breaking} introduce synthetic ASR errors and ASR related features into the source side of MT corpora. 
\item Techniques such as domain adaptation~\citep{liu2003use,fugen2008system}, re-segmention~\citep{matusov2006automatic}, punctuation restoration~\citep{fugen2008system}, disfluency detection~\citep{fitzgerald2009reconstructing,wang2018semi,dong2019adapting} and so on, are proposed to provide the translation model with well-formed and domain matched text inputs.
\item Many efforts turn to strengthen the tight integration between the ASR output and the MT input, such as the n-best translation, lattices and confusion nets and so on~\citep{sperber2020speech}. 
\end{inparaenum} 

On the other hand, a paradigm shift towards end-to-end system is emerging to alleviate the drawbacks of cascaded systems.
\citet{berard2016listen,duong2016attentional} have given the first proof that it's promising to solve speech-to-text translation in an end-to-end way. After that, end-to-end ST has attracted many attentions in industry and research communities~\citep{vila2018end,salesky2018towards,salesky2019fluent,di2019adapting,bahar2019comparative,di2019enhancing,inaguma2020espnet}. 
Prior work mainly explores pre-training with ASR and MT tasks to get a good point of initialization for ST models~\citep{weiss2017sequence,berard2018end,bansal2018pre,stoian2020analyzing}. Additionally, multi-task learning can efficiently transfer knowledge between different tasks~\citep{anastasopoulos2018tied,vydana2020jointly}. Pre-training and multi-task learning have also been widely used in recent work~\citep{wang2020bridging,dong2020listen}. 
Since there still exists a great gap for the performance between cascaded and end-to-end ST systems, many methods have been proposed to improve the end-to-end ST models, such as curriculum learning~\citep{kano2018structured,wang2020curriculum}, knowledge distillation~\citep{liu2019end}, modality agnostic meta-learning~\citep{indurthi2019data}, two-pass decoding~\citep{sung2019towards}, attention-passing~\citep{sperber2019attention}, data augmentation~\citep{jia2019leveraging,bahar2019using,pino2019harnessing}, model adaptation~\citep{di2020instance}, self-training~\citep{pino2020self} and so on. 

Due to the scarcity of data resource, how to efficiently utilize ASR and MT parallel data is a big problem for ST, especially in the end-to-end situation. However, existing end-to-end methods mostly resort to ordinary pretraining or multitask learning to integrate external ASR resources, which may face the issue of catastrophic forgetting and modal mismatch. And it is still challenging for previous methods to leverage external bilingual MT data efficiently.

\section{Proposed \method Approach}
\label{sec:approach}

\subsection{Overview}

The detailed framework of our method is shown in Figure~\ref{fig:overview}. To be specific, the speech translation model accepts the original audio feature as input and outputs the target text sequence. We divide our method into two phases, including the acoustic-semantic modeling phase (AS) and the transcription-translation modeling phase (TT). 
Firstly, the AS phase accepts the speech features, outputs the acoustic representation, and encodes the shrunk acoustic representation into semantic representation. 
In this work, the small-grained unit, phonemes are selected as the acoustic modeling unit. 
Then, the TT phase accepts the AS's representation and consecutively outputs source transcription and target translation text sequences with a single shared  decoder.

\paragraph{Problem Formulation} The speech translation corpus usually contains speech-transcription-translation triples. We add phoneme sequences to make up quadruples, denoted as $\mathcal{S}=\{(\xx,\uu,\zz,\yy)\}$ (More details about the data preparation can be seen in the experimental settings). Specially, $\xx=(x_1,...,x_{T_x})$ is a sequence of acoustic features. $\uu=(u_1,...,u_{T_{u}})$, $\zz=(z_1,...,z_{T_{z}})$, and $\yy=(y_1,...,y_{T_{y}})$ represents the corresponding phoneme sequence in source language, transcription in source language and the translation in target language respectively. Meanwhile, $\mathcal{A}=\{(\zz',\yy')\}$ represents the external text translation corpus, which can be utilized for pre-training the decoder. Usually, the amount of end-to-end speech translation corpus is much smaller than that of text translation, i.e. $|\mathcal{S}| \ll |\mathcal{A}|$. 

\subsection{Acoustic-Semantic Modeling}
\label{sec:sa}
The acoustic-semantic modeling phase takes the input of low-level audio features $\xx$ and outputs a series of vectors $\hh_\text{AS}$ corresponding to the phoneme sequence $\uu$ in the source language. 
Different from the general sequence-to-sequence models, two modifications are introduced. 
Firstly, in order to preserve more acoustic information, we introduce the supervision signal of the connectionist temporal classification (CTC) loss function, a scalable, end-to-end approach to monotonic sequence transduction~\citep{graves2006connectionist,salazar2019self}.
Secondly, since the length of audio features is much larger than that of source phoneme ($T_x\gg T_u$),  we introduce a shrinking method which can skip the blank-dominated steps to reduce the encoded sequence length.
\paragraph{Self-Attention with CTC} General preprocessing includes down-sampling and linear layers. Down-sampling refers to the dimensionality reduction processing of the input audio features in the time and frequency domains. 
In order to simplify the network, we adopt manual dimensionality reduction, that is, a method of sampling one frame every three frames. The linear layer maps the length of the frequency domain feature of the audio feature to the preset network hidden layer size. After preprocessing, multiple Transformer blocks are stacked for acoustic feature extraction.

\begin{equation}
     \hat \hh_{AS} = \text{Attention}(\text{Linear}(\text{Down-sample}(\textbf{x})))
\end{equation}
Finally, the softmax operator is applied to the result of the affine transformation to obtain the probability of the phoneme sequence. CTC loss is adopted to accelerate the convergence of acoustic modeling. 
CTC assumes $T_u \leq T_x$, and defines an intermediate alphabet $\mathcal{V}'=\mathcal{V}\cup{\{blank\}}$. A \textit{path} $\pi$ is defined as a $T_x$-length sequence of intermediate labels $\pi = (\pi_1,...,\pi_{T_x}) \in \mathcal{V}^{'T_x}$. And a many-to-one mapping is defined from paths to output sequences by removing blank symbols and consecutively repeated labels. 

The conditional probability of a given labelling $\uu$ $\in\mathcal{V}^{'T_u}$ can be modeled by marginalizing over all paths corresponding to it.
The distribution over the set $\mathcal{V}^{'T_x}$ of \textit{path} $\pi$ is defined by the probability of a sequence of conditionally-independent outputs, which can be calculated non-autoregressively.
\begin{equation}
p_{ctc} (\uu|\xx) = \sum_{\pi \in\mathcal{B}^{-1}(\textbf{u})}p(\pi| \hat\hh_{AS})
\end{equation}

\begin{equation}
p(\pi| \hat\hh_{AS}) = \prod_{t=1}^{T_x} p(\pi_{t}|\hat \hh_{AS})
\end{equation}

And $p(\pi_{t}|\hat \hh_{AS})$ is computed by applying the $softmax$ function to $logits$.
Finally, the objective training function during AS phase is defined as:
\begin{equation}
    \mathcal{L}_{\text{AS}} = - \log p_{ctc} (\uu|\xx)
    \label{ctc_loss}
\end{equation}

\paragraph{Acoustic Unit Shrinking} 
The shrinking layer aims at reducing the potential blank frames, and repeated frames.
The details can be seen in the sub-figure of Figure \ref{fig:overview}. 
The method is mainly founded on the studies of \citet{chen2016phone,yi2019ectc}. We adopt the implementation by removing the blank frames and averaging the repeated frames. Without the interruption of blank and repeated frames, the language modeling ability should be better in theory. Blank frames can be detected according to the spike characteristics of CTC probability distribution.
\begin{equation}
    \hh_{AS}' = \text{Shrink} (\hat \hh_{AS}, p_{ctc}(\uu|\xx))
\end{equation}
Then, similarly, after shrinking, multiple Transformer blocks are stacked to extract higher-level semantic representations and result in the final output $\hh_{AS}$.
\begin{equation}
    \hh_{AS} = \text{Attention}(\hh_{AS}')
\end{equation}
\subsection{Transcription-Translation Modeling}
\label{sec:att}
We jointly model the transcription and translation generation in a single  shared decoder, which takes the acoustic representation $\hh_{AS}$ as the input and generates the source text $\zz$ and target text $\yy$.
This TT phase is stacked with $T$ Transformer blocks, consisting of multi-head attention layers and feed-forward networks. 
\begin{equation}
    \hh_{TT} = \text{Transformer}([\zz,\yy], \hh_{AS})
\end{equation}
As shown in Figure~\ref{fig:overview}, the decoder output is the tandem result of the transcription and translation sequences, joined by the task identificator token (``$<$asr$>$" for recognition and ``$<$st$>$" for translation), marked as $[\zz,\yy]$. That is to say, the model is able to continuously predict the transcription sequence and the translation sequence. The training objective of the TT phase is the cross entropy between prediction sequence and target sequence.
\begin{equation}
    \mathcal{L}_{\text{TT}} = - \log p ([\zz,\yy]|\xx)
    \label{ce_loss}
\end{equation}
Compared with the multi-task learning method, consecutive decoding can make prediction from easy (transcription) to hard (translation) tasks, alleviating the decoding pressure. For example, when predicting the translation sequence, since the corresponding transliteration sequence has been decoded, that is, the intermediate recognition result of the known speech translation and the source of information for decoding, the translation sequence can be improved. 

\paragraph{Pre-train the Consecutive Decoder} Generally, it is straightforward to use ASR corpus to improve the performance of ST systems, but is non-trivial to utilize MT corpus. Taking advantage of the structure of consecutive decoding, we propose a method to enhance the performance of ST systems by means of external MT paired data. Inspired by translation language modeling (TLM) in XLM~\citep{lample2019cross}, we use a masked loss function to pre-train TT phase. Specifically, we use external data in $\mathcal{A}$ to pre-train the parameters of the TT part. Different from the end-to-end training stage, there is no audio feature as input during pre-training, so cross-attention cannot attend to the output of the previous AS phase. We use an all-zero constant, marked as $\hh_{AS_{blank}}$ to substitute the encoded representations ($\hh_{AS}$) from TT phase to be consistent with fine-tuning. When calculating the objective function, we mask the loss for prediction of the recognition result, and make the decoder predicts the translation sequence when aware of the input of the transcription sequence. The translation loss of the TT phase during pre-training only includes the masked cross entropy:

\begin{equation}
    \mathcal{L}_{\text{TT}_{\text{PT}}} = - \sum _{i=1}^{T_y} \log p(y_i|\zz, y_{< i})
    \label{mt_loss}
\end{equation}

\subsection{Joint Learning}
\label{subsec:joint}
Apart from pre-training, we exploit joint learning to integrate our unified ST model. The total training objective is as follows:
\begin{equation}
    \mathcal{L} = \alpha \mathcal{L}_{\text{AS}} + (1-\alpha) \mathcal{L}_{\text{TT}}
    \label{joint_loss}
\end{equation}
Where $\alpha$ is a tunable parameter to balance the objectives of different phases.

\begin{algorithm}
\caption{\method without pre-training}
\label{algorithm_scratch}
\begin{algorithmic}[1] 
    \State \# training from scratch ($\theta_{AS}^0 \rightarrow \theta_{AS}^1, \theta_{TT}^0 \rightarrow \theta_{TT}^1$)
    \While{not converged}
        \State supervised training ST with $(\xx,\uu,\yy) \in \mathcal{S}$
    \EndWhile
    \State \textbf{return} ST with $\theta_{AS}^1, \theta_{TT}^1$
\end{algorithmic}
\end{algorithm}
\begin{algorithm}
\caption{\method with pre-training}
\label{algorithm_pt}
\begin{algorithmic}[1] 
    \State \# pre-training ConDec ($\theta_{AS}^0 \rightarrow \theta_{AS}^0, \theta_{TT}^0 \rightarrow \theta_{TT}^1$)
    \While{not converged}
        \State CE loss guided supervised training ConDec with $(\zz',\yy') \in \mathcal{A}$
    \EndWhile
    \State \# pre-training AM ($\theta_{AS}^0 \rightarrow \theta_{AS}^1, \theta_{TT}^1 \rightarrow \theta_{TT}^1$)
    \While{not converged}
        \State CTC loss guided supervised training AM with $(\xx, \uu) \in \mathcal{S}$
    \EndWhile
    \State \# fine-tuning ST ($\theta_{AS}^1 \rightarrow \theta_{AS}^2, \theta_{TT}^1 \rightarrow \theta_{TT}^2$)
    \While{not converged}
        \State Supervised training ST with $(\xx, \uu, \zz, \yy) \in \mathcal{S}$
    \EndWhile
    \State \textbf{return} ST with $\theta_{AS}^2, \theta_{TT}^2$
\end{algorithmic}
\end{algorithm}

We design different training algorithms for our method training from scratch (seen in Algorithm \ref{algorithm_scratch}) and training with pre-training the consecutive decoder (seen in Algorithm \ref{algorithm_pt}). 
Algorithm \ref{algorithm_pt} is determined after many attempts to better avoid the catastrophic forgetting of pre-trained knowledge.

\section{Experiments}
\label{sec:exps}

\begin{table*}[ht]
\centering
\small
\begin{tabular}{lccc} 
\toprule
 Method & \shortstack{Enc Pre-train\\(speech data)} & \shortstack{Dec Pre-train\\(text data)}  & BLEU \\  
\midrule
\textbf{MT system} & & &    \\
Transformer MT~\citep{liu2019end} & - & - & 22.91 \\
\midrule
\textbf{Base setting} & & &    \\
LSTM ST~\citep{berard2018end} & \xmark & \xmark&   12.90 \\
~~+pre-train+multitask~\citep{berard2018end} & \cmark & \cmark  & 13.40 \\
LSTM ST+pre-train~\citep{inaguma2020espnet} & \cmark & \cmark    &16.68\\
Transformer+pre-train~\citep{liu2019end} & \cmark & \cmark  &14.30\\
~~+knowledge distillation~\citep{liu2019end} &\cmark &\cmark  &17.02\\
TCEN-LSTM~\citep{wang2020bridging}& \cmark & \cmark  &17.05\\
Transformer+ASR pre-train~\citep{wang2020curriculum} &\cmark & \xmark  &15.97\\
~~+curriculum pre-train~\citep{wang2020curriculum} &\cmark & \xmark &17.66\\
LUT~\citep{dong2020listen} &\xmark & \xmark & 17.75 \\
\method without pre-training &\xmark & \xmark  &\textbf{17.83}\\
\midrule 
\textbf{Expanded setting}& &  &  \\
LSTM+pre-train+SpecAugment~\citep{bahar2019using} & \cmark (236h)&\cmark &17.00\\
Multi-task+pre-train~\citep{inaguma2019multilingual}  & \cmark(472h)&\xmark &17.60\\
Transformer+ASR pre-train~\citep{wang2020curriculum} &\cmark(960h) &\xmark &16.90\\
~~+curriculum pre-train~\citep{wang2020curriculum} &\cmark(960h) &\xmark &18.01\\
LUT~\citep{dong2020listen} &\cmark(207h) & \xmark&\textbf{18.34} \\
\method with pre-training & \cmark & \cmark(1M) & 18.23\\
\bottomrule
\end{tabular}
\caption{Performance on Augmented Librispeech English-French test set. \method achieves the best performance compared with previous work.}
\label{enfr}
\end{table*}

\begin{table}[]
    \centering
    \begin{tabular}{lccc}
    \toprule
    & \#train & \#dev & \#test\\
    \midrule
   Augmented LibriSpeech  & 47,271 & 1,071& 2,048\\
   IWSLT2018 English-German  & 171,121 & 653 & 793\\
   TED English-Chinese  & 308,660 & 835 & 1,223\\
    \bottomrule
    \end{tabular}
    \caption{Statistics of utterances for the three ST datasets.}
    \label{tab:datasets}
\end{table}

\begin{table*}[ht]
\centering
\begin{tabular}{p{7.8cm}p{1.5cm}<{\centering}p{1.5cm}<{\centering}p{0.8cm}<{\centering}} 
\toprule
 Method & \shortstack{Enc Pre-train\\(speech data)} & \shortstack{Dec Pre-train\\(text data)} &  tst2013 \\
\midrule
\textbf{MT system} & & &    \\
RNN MT~\citep{inaguma2020espnet} & - & - & 24.90\\
\midrule
\textbf{Base setting} & & & \\
ESPnet~\citep{inaguma2020espnet} &\xmark &\xmark &  12.50\\
~~+enc pre-train &\cmark &\xmark &  13.12\\
~~+enc dec pre-train&\cmark &\cmark &  13.54\\
Transformer+ASR pre-train~\citep{wang2020curriculum}&\cmark & \xmark& 15.35\\
~~+curriculum pre-train~\citep{wang2020curriculum}&\cmark &\xmark & 16.27 \\
LUT~\citep{dong2020listen} & \xmark & \xmark&    \textbf{16.35} \\
\method without pre-training &\xmark &\xmark & 16.30 \\
\midrule 
\textbf{Expanded  setting}& & &  \\
Multi-task+pre-train ~\citep{inaguma2019multilingual}&\cmark(472h) &\xmark & 14.60 \\
CL-fast*~\citep{kano2018structured}&\cmark(479h) &\xmark &14.33 \\
TCEN-LSTM~\citep{wang2020bridging} &\cmark(479h) &\cmark(40M) &   17.67\\
Transformer+curriculum pre-train~\citep{wang2020curriculum} &\cmark(479h) &\cmark(4M) & 18.15\\
LUT~\citep{dong2020listen} &\cmark(207h) & \xmark&   18.59\\
\method with pre-training & \cmark &\cmark(1M) & \textbf{18.63} \\
\bottomrule
\end{tabular}
\caption{Performance on English-German TED test set. *: re-implemented~\cite{wang2020curriculum}. \method achieves the best performance compared with previous work.}
\label{ende}
\end{table*}

\subsection{Dataset and Preprocessing}
Our experiments are conducted on three open sourced datasets belonging to three different language pairs).
The following is a detailed description of the three datasets.
Table \ref{tab:datasets} shows the statistics of utterances for the three datasets. 

\paragraph{Augmented LibriSpeech Dataset}
Augmented LibriSpeech is an expanded version from the ASR dataset, LibriSpeech, by automatically aligning e-books in different languages, and the translation reference is doubled via Google Translate.

\paragraph{IWSLT2018 English-German Dataset}
IWSLT2018 English-German is the officially used dataset for the open evaluation campaign on spoken language translation.
We utilize the attached timestamps to segment a raw long audio into chunks and remove samples missing the target language translation. 
dev2010 and tst2013 are used as validation set and test set.

\paragraph{TED English-Chinese Dataset}
TED English-Chinese Dataset~\cite{liu2019end} is the first released dataset for end-to-end speech translation from English to Chinese. We use dev2010 as development set and tst2015 as test set. 

\paragraph{WMT Machine Translation Corpus} 
We use WMT14\footnote{\url{https://www.statmt.org/wmt14/translation-task.html}} English-to-French and English-to-German training data as the external MT parallel corpus ($\in\mathcal{A}$) in the expanded experimental setting for reproducibility. 
We pre-processed all of the data of specific language pairs, and filtered sentence pairs whose total length exceeds 500. We shuffled the data and randomly selected a subset of 1 million for the following experiments and analysis.

\subsection{Experimental Setup}

Our acoustic features are 80-dimensional log-Mel filterbanks extracted with a step size of 10ms and window size of 25ms and extended with mean sub-traction and variance normalization. The features are stacked with 5 frames to the right. 
For all source language text data, we lower case all the texts, tokenize and remove the punctuation to make the data more consistent with the output of ASR.
For target French and German text data, we lower case all the texts, tokenize and apply normalize punctuations with the Moses scripts\footnote{\url{https://github.com/moses-smt/mosesdecoder}}.
For target Chinese text data, we use the raw released segmented results.
For English-French and English-German datasets, we apply BPE\footnote{\url{https://github.com/rsennrich/subword-nmt}}~\citep{sennrich2016neural} to the combination of source and target text to obtain shared subword units. And for English-Chinese dataset, we apply BPE to the source text and target text respectively.
The number of merge operations in BPE is set to $8$k for all datasets. 
In order to simplify, we use the open-source grapheme to phoneme tool\footnote{\url{https://github.com/Kyubyong/g2p}} to map the transcription to the phoneme sequence (An example in Table \ref{tab:quadruple}). 
The alphabet of labels $\mathcal{V}$ includes the union of sub-word vocabulary and phoneme vocabulary, plus a few special symbols (including ``$<$asr$>$", ``$<$st$>$" and ``blank"). 
For English-French and English-German translation, we evaluate the accuracy by case-insensitive BLEU score using the \texttt{multi-bleu.pl}\footnote{\url{https://github.com/moses-smt/mosesdecoder/scripts/generic/multi-bleu.perl}} script. While for English-Chinese translation, we report character-level BLEU score.
We use word error rates (WER) and phoneme error rates (PER) to evaluate the prediction accuracy of transcription and phoneme sequences, respectively.

\begin{table}[h]
    \small
    \begin{tabular}{p{0.95\linewidth}}
    \toprule
       \textbf{speech}  \quad  135-19215-0118.wav\\
       \textbf{phonemes} \\
       \quad Y UW1 $<$space$>$ M AH1 S T $<$space$>$ M EY1 K $<$space$>$ AH0 $<$space$>$ D R IY1 M $<$space$>$ W ER1 L $<$space$>$ ER0 AW1 N D $<$space$>$ DH AH0 $<$space$>$ B R AY1 D\\
       \textbf{transcription}  \\
       \quad you must make a dream whirl around the bride\\
       \textbf{translation} \\
       \quad il faudrait faire tourbillonner un songe autour de l' épousée .\\
    \bottomrule
    \end{tabular}
    \caption{An example of the speech-phoneme-transcription-translation quadruples. 
    Phonemes can be converted from the transcription text.  
    }
    \label{tab:quadruple}
\end{table}

We use a similar hyperparameter setting with the base Transformer model~\citep{vaswani2017attention}. 
The number of transformer blocks is set to 12 and 6 for the acoustic-semantic (AS) phase and the transcription-translation (TT) phase, respectively.
And phoneme supervision is added to the middle layer of AS phase for all datasets.
SpecAugment strategy~\citep{park2019specaugment} is adopted to avoid overfitting with frequency masking (F = 30, mF = 2) and time masking (T = 40, mT = 2).
All samples are batched together with 20000-frame features by an approximate feature sequence length during training. We train our models on 1 NVIDIA V100 GPU with a maximum number of $400$k training steps. We use the greedy search decoding strategy for our experimental settings. The maximum decoding length is set to 500 for our models with consecutive decoding and 250 for other methods on all datasets.
$\alpha$ in Equation \ref{joint_loss} is set to $0.5$ for all datasets (We have searched the value of $\alpha$ using a step of 0.2.).
We design different workflows for our method training from scratch and training with pre-training the consecutive decoder. 
More details are in the results. The final model is averaged on the last 10 checkpoints.

\section{Results}
\label{sec:res}

\begin{table*}[ht]
\centering
\small
\begin{tabular}{lccc} 
\toprule
 Method & \shortstack{Enc Pre-train\\(speech data)} & \shortstack{Dec Pre-train\\(text data)}  & BLEU \\  
\midrule
\textbf{MT system} & & &    \\
Transformer MT~\citep{liu2019end} & - & - & 27.08\\ 
\midrule
\textbf{Base setting} & & &    \\
Transformer+pre-train~\citep{liu2019end} & \cmark & \cmark & 16.80 \\
~~+knowledge distillation~\citep{liu2019end}& \cmark &\cmark & 19.55 \\
Multi-task+pre-train*~\citep{inaguma2019multilingual}(re-implemented) & \cmark & \xmark& 20.45 \\
LUT~\citep{dong2020listen} & \xmark& \xmark  & 20.84\\
\method without pre-training & \xmark& \xmark  &\textbf{21.12}\\
\bottomrule
\end{tabular}
\caption{Performance on English-Chinese TED test set. \method achieves the best performance compared with previous work.}
\label{enzh}
\end{table*}

\subsection{Baselines}
We compare with systems in different settings:
\paragraph{Base setting:}
ST models are trained with only ST triple corpus.
\paragraph{Expanded setting:}
ST models are trained with ST triple corpus augmented with external ASR and MT corpus. 
In the context of expanded setting, ~\citet{bahar2019using} apply the SpecAugment~\citep{park2019specaugment} with a total of 236 hours of speech for ASR pre-training. ~\citet{inaguma2019multilingual} combine three ST datasets of 472 hours training data to train a multilingual ST model. \citet{wang2020bridging} introduce an additional 207 hours ASR corpus and 40M parallel data from WMT18 to enhance the ST. 
We mainly explored additional MT data in this work.
\paragraph{MT system:}
Text translation models are trained with manual transcribed transcription-translation pairs, which can be regarded as the upper bound of speech translation tasks.

\subsection{Main Results}
\begin{table}[ht]
\centering
\small
\begin{tabular}{cll} 
\toprule
 &Method & BLEU \\  
\midrule
\multirow{2}*{\shortstack{$En$$\rightarrow$$Fr$}} & Pipeline & 17.58 \\
&\method & 17.83\\
\midrule
\multirow{2}*{\shortstack{$En$$\rightarrow$$De$}} & Pipeline & 15.38 \\
&\method & 16.30\\
\midrule
\multirow{2}*{\shortstack{$En$$\rightarrow$$Zh$}} & Pipeline & 21.36 \\
&\method & 21.12\\
\bottomrule
\end{tabular}
\caption{\method versus cascaded systems on three datasets. ``Pipeline" means cascading independent ASR and MT models.}
\label{cascaded}
\end{table}

We conduct experiments on three public datasets.

\paragraph{Librispeech English-French}
In Table \ref{enfr}, we compared the performance with previous end-to-end methods for En-Fr dataset. \method outscored or on par with the previous best results both in the base setting and the expanded setting. 
We achieved better results than a knowledge distillation baseline in which the ST model can learn from an strong MT model~\citep{liu2019end}. 
We also exceeded~\citet{wang2020bridging, wang2020curriculum}, even though they used more external ASR and MT data. Although our work is not significantly superior to \citet{dong2020listen}, \method can make full use of the machine translation corpus compared with previous work. 
With an additional 1 million sentence pairs, we achieve +0.4 BLEU score improvements (17.83 v.s. 18.23). This proposal promises great potential for the application of the \method. 

\paragraph{IWSLT2018 English-German}
In Table \ref{ende}, we compared the performance with previous end-to-end methods for En-De dataset, which is a more difficult situation for speech translation.
Although the performance of ST model is far inferior to that of MT model, our method had a slight advantage as compared to previous competitors on tst2013 in the expanded setting. 

\paragraph{TED English-Chinese}
In Table \ref{enzh}, we compared the performance with previous end-to-end methods for En-Zh dataset. Our work outperformed the previous best methods obviously by about 0.3 BLEU in the base setting. \method exceeded \citet{dong2020listen} augmented by knowledge distillation from BERT~\citep{devlin2018bert} with a big margin.

\paragraph{Comparison with Cascaded Systems}~In Table \ref{cascaded}, we compare the performance of our E2E models with the cascaded systems. It shows that E2E models are outstanding or comparable on all En$\rightarrow$Fr/De/Zh tasks, proving our method's capacity for joint optimization of the separate ASR and MT tasks in a model. 

\subsection{Ablation Study}
We use an ablation study to evaluate the importance of different modules in our methods. The results in Table~\ref{ablation} show that all the methods adapted are positive for the model performance, and the benefits of different parts can be superimposed. Models with consecutive decoding are able to predict both the recognition and translation, for which we also report WER and PER to evaluate the performance of different modeling phase. It has been proved that consecutive decoding brings the gain of 1 BLEU compared with the base model and pre-training decoder can bring improvements to all three metrics. 

\begin{table}[ht]
\centering
\small
\begin{tabular}{lccc}   
\toprule
 & BLEU$\uparrow$ & WER$\downarrow$ & PER$\downarrow$ \\  
\midrule        
  \method & \textbf{18.23} & \textbf{14.60} & \textbf{10.30}\\
  ~~~ w/o PT Dec  & 17.51  & 15.30 & 11.90\\
  ~~~ w/o CD  & 16.57 & - & -\\
  ~~~ w/o Shrink  & 16.40 & - & -\\
 ~~~ w/o AS loss *  & 15.48 & - &-\\
  ~~~ w/o AS loss   & 11.24 & - &-\\
\bottomrule
\end{tabular}
\caption{
Benefits of each component in \method on En-Fr test set. 
``PT Dec" stands for pre-training the successive decoder.
``CD" represents using the consecutive decoder.
``*" means using ASR pre-training as initialization.
}
\label{ablation}
\end{table} 

\begin{table*}[h]
\centering
\begin{tabular}{ccccccccccc} 
\toprule
Error Range & 0 & 1 & 2 & 3 & 4 & 5 & 6 & 7 & 8 & 9 \\
Probability & 0.32 & 0.66 & 0.83 & 0.91 & 0.95 & 0.97 & 0.98 & 0.99 & 0.99 & 0.99 \\
\bottomrule
\end{tabular}
\caption{Statistics of the absolute error between the length of shrunk acoustic unit and the length of the gold phoneme sequence.}
\label{table:shrink}
\end{table*}

\subsection{Effects of Pre-training}
\begin{figure}[ht]
    \centering
    \includegraphics[width=0.50\textwidth]{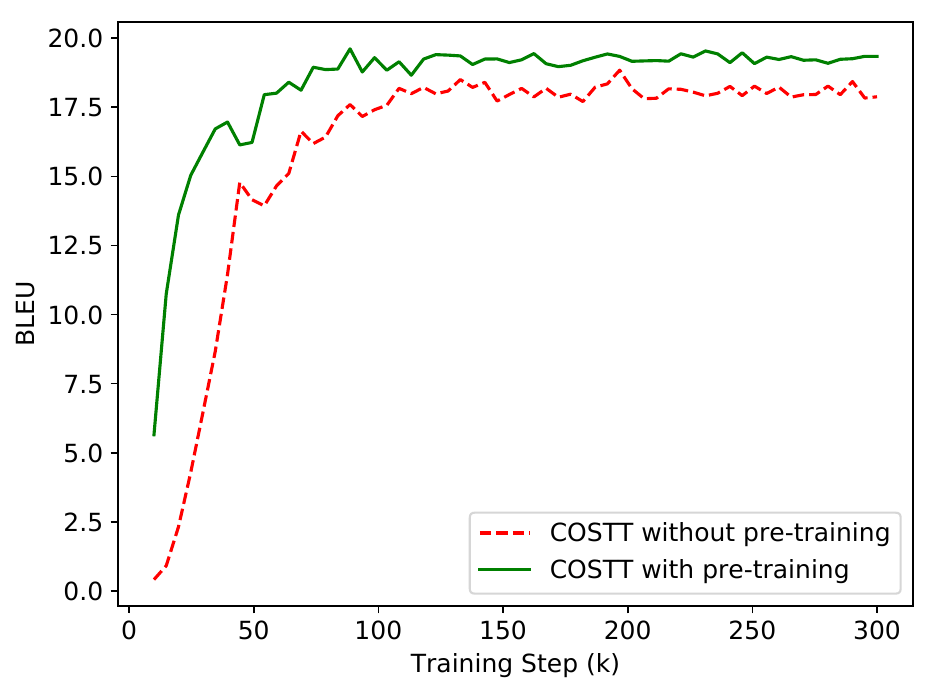}
    \caption{BLEU scores on Augmented Librispeech validation set for \method with and without pre-training on extra parallel MT corpus. Notice that the full \method with MT pre-training does improve the performance. }
    \label{fig:work_flow}
\end{figure}

Figure \ref{fig:work_flow} shows the convergence curve on the English-French validation set of the two training algorithms in Algorithm~\ref{algorithm_scratch} and  Algorithm~\ref{algorithm_pt}.. It proves that \method with pre-training the consecutive decoder can get a better initialization and converge better benefiting from our flexible model structure. 

\subsection{Parameters of ST systems}
The parameter sizes of different systems are shown in Table~\ref{model_size}. The pipeline system needs a separate ASR model and MT model, so its parameters are doubled. Our method \method only needs the same parameters as the vanilla end-to-end model, but it can achieve superior performance thanks to the consecutive decoding mechanism.

\begin{table}[ht]
\centering
\begin{tabular}{lccc} 
\toprule
Model & Params\\
\hline
Pipeline & 110M\\
E2E & 55M & & \\
\method (12 L) & 55M\\
\method (18 L) & 76M \\
\bottomrule
\end{tabular}
\caption{Statistics of parameters of different ST systems. E2E: the vanilla end-to-end ST system.}
\label{model_size}
\end{table}

\subsection{Effects of Shrinking Mechanism}

\begin{figure}[!h]
    \centering
    \includegraphics[width=0.45\textwidth]{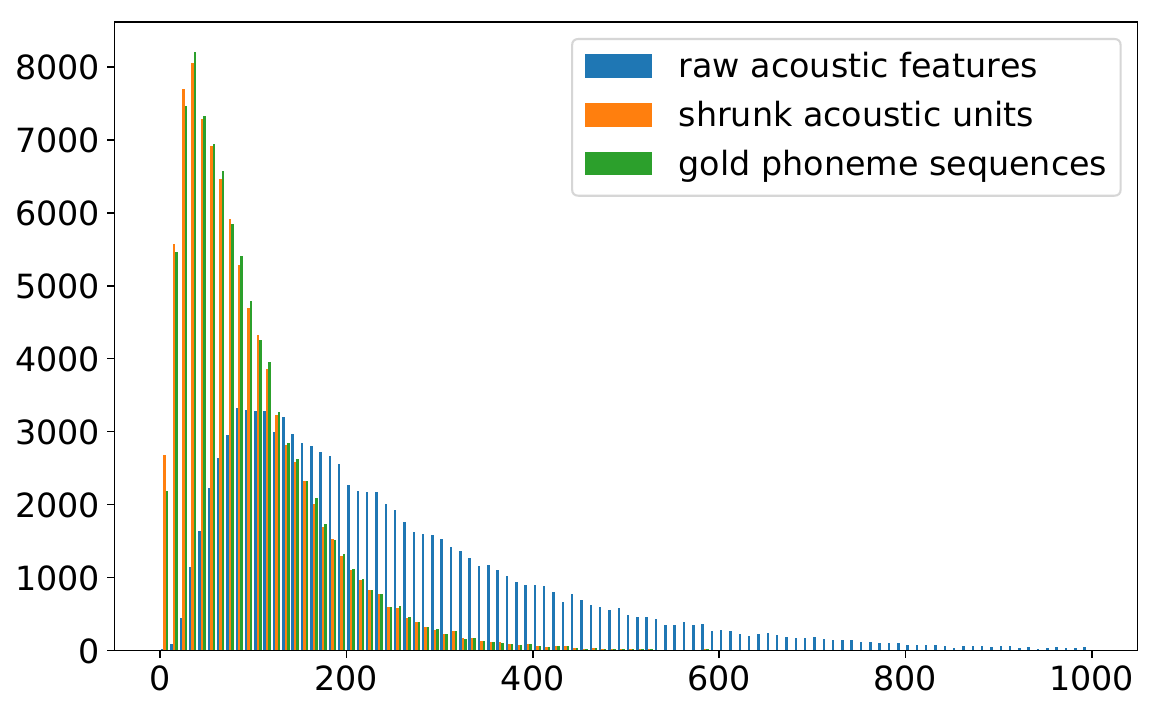}
    \caption{Length distribution of the raw acoustic features, the shrunk acoustic units and the gold phoneme sequences on English-French training set}
    \label{fig:histogram}
\end{figure}

In order to verify whether the shrinking mechanism has achieved the expected effect, we collected the sequence length of the encoded hidden layer before and after shrinking and the length distribution of the gold phoneme sequence. As shown in Figure \ref{fig:histogram}, the sequence length of the shrunk acoustic unit and the distribution of phoneme length are almost the same. 
According to statistics in Table~\ref{table:shrink}, for more than 90\% of the samples, the absolute error between the length of the shrunk acoustic unit and the length of the gold phoneme sequence is within 3. 
Moreover, the length of the shrunk acoustic unit is significantly reduced compared to the length of the original acoustic feature. 
The results show that the shrinking mechanism can detect blank frames and repeated frames well, while reducing the computational resources and preventing memory overflow.

\subsection{Effects of Layers after Shrinking}

As mentioned in previous sections, our model stacks additional Transformer blocks after the shrinking operation. We have conducted simplified experiments on the English-French dataset with a vanilla speech translation model without consecutive decoding to demonstrate the importance of the additional encoding layers after shrinking. The output of the encoded layer uses the CTC loss as the supervision, and we use the subword of transcriptions in the source language as the acoustic labels. Results can be seen in Table~\ref{tab:semanticencoder}. The experimental results show that directly inputting the shrunk encoded output to the decoder will cause performance loss. And stacking additional encoding layers after shrinking can bring significant performance improvements. We conjecture that there is a lack of semantic encoding modules between acoustic encoding and linguistic decoding. In addition, the relationship between the hidden states after shrinking has changed a lot, and an additional network structure is required to re-extract high-level encoded features.

\begin{table}[h]
    \centering
    \begin{tabular}{cccc}
    \toprule
      Encoder & Shrinking & Layers after Shrinking  & BLEU \\
     \midrule
      6 & \xmark & - & 12.70\\
      6 & \cmark & - & 11.34\\
      6 & \cmark & 6 & 16.46 \\
    \bottomrule
    \end{tabular}
    \caption{BLEU scores on Augmented Librispeech test set for different model configurations. The number represents the layers of Transformer block contained in the corresponding module.}
    \label{tab:semanticencoder}
\end{table}

\subsection{Case Study on English-French}
The cases in Table \ref{Case} shows that \method has obvious structural advantages in solving missed translation, mistranslation, and fault tolerance. For instance: \#1, the base model missed the translation of ``yes" in the audio, whereas our method produced a completely correct translation.  After listening to the original audio, it is suspected that the missing translation is due to an unusual pause between ``doctor" and ``yes". 
\#2, the base model mistranslated the ``aboard" in the audio into ``vers l’ avant"(``forward" in English), yet our method could correctly translate it into  ``a bord" based on the correct transcription prediction. The reason for the mistranslation may be that the audio clips are pronounced similarly, thus confusing the translation model.  
\#3, the base model translated erroneously most of the content, and our model also predicted ``today" in the audio as ``to day". However, in the end, our method was able to predict the translation result completely and correctly.

\begin{table}[h]
    \centering
    \small
    \begin{tabular}{lp{2.25in}}
    \toprule
      Speech \#1 & 766-144485-0043.wav\\
      Transcript & said the doctor yes \\
      Target & dit le docteur \textbf{, oui} .\\
      Base ST & dit le docteur .\\
      \method    & \textit{$<$asr$>$ said the doctor \underline{yes}} $<$ast$>$ dit le docteur \textbf{, oui} . \\
      \midrule
      Speech \#2 & 2488-36617-0066.wav\\
      Transcript & i rushed aboard\\
      Target & je me précipitai à bord.\\
      Base ST & je me précipitai \textbf{vers l' avant} .\\
      \method   & \textit{$<$asr$>$ i rushed \underline{aboard}} $<$ast$>$ je me précipitai \textbf{à bord} . \\
      \midrule
      Speech \#3 & 766-144485-0098.wav\\
      Transcript & is there any news today\\
      Target & \textbf{y a-t-il des nouvelles} aujourd' hui ?\\
      Base ST & \textbf{est-ce que j' ai déjà utilisé} aujourd' hui ?\\
      \method   & \textit{$<$asr$>$ is there any news \underline{to day}} $<$ast$>$ \textbf{y a-t-il des nouvelles} aujourd' hui ?\\
     \bottomrule
    \end{tabular}
    \caption{
    Examples of speech translation generated by \method and the baseline ST model.  Words in bold highlight the difference. 
    Words underlined, as generated by \method, contributes to the improved translation results.}
    \label{Case}
\end{table}

\subsection{Compared with 3-stage Pipeline}
In the case study of Table ~\ref{Case}, we have listed some examples of errors in transcription recognition, but COSTT can still correctly predict the translation sequence, which proves that COSTT can solve the error propagation problem to some extent.
In a pipeline system that includes the phoneme stage, the phoneme recognition error will also lead to error propagation.
But in \method, the phoneme sequence is only the intermediate supervision used during training, not necessary during inferring. Moreover, end-to-end training can alleviate the error propagation between different stages.
We believe that the more stages, the greater the advantage of our method.
We have built a 3-stage system consisting of acoustics-to-phoneme (A2P), phoneme-to-transcript (P2T), and transcript-to-translation (T2T) stages. A2P is a phoneme recognition model based on the CTC loss function, which uses phoneme error rate (PER) to evaluate performance (the lower the better). Both P2T and T2T use the sequence-to-sequence model based on Transformer and BLEU is the evaluation criterion (the higher the better).
The performance of each module is shown in Table~\ref{tab:3-stage}.
The performance of different systems in Table~\ref{tab:3-stage-pipeline} prove that with the increase of stages, the problem of error propagation becomes more and more serious, which shows the benefits of the \method method.

\begin{table}[h]
    \centering
    \begin{tabular}{cccc}
    \toprule
      Stage & A2P (PER) & P2T (BLEU) & T2T (BLEU) \\
      Performance & 10.30 & 92.08 & 21.51 \\
    \bottomrule
    \end{tabular}
    \caption{Performance of each module of our 3-stage Pipeline.}
    \label{tab:3-stage}
\end{table}

\begin{table}[h]
    \centering
    \begin{tabular}{cc}
    \toprule
      System & BLEU \\
      \midrule
      3-stage Pipeline  & 12.22\\
      2-stage Pipeline  & 17.58\\
      COSTT   & \textbf{18.23}\\
    \bottomrule
    \end{tabular}
    \caption{\method versus 2-stage Pipeline and 3-stage Pipeline on Augmentated Librispeech En-Fr test set.}
    \label{tab:3-stage-pipeline}
\end{table}

\section{Conclusion}
\label{sec:conclusion}

We propose \method, a novel and unified training framework for jointly speech recognition and speech translation. We use the consecutive decoding strategy to realize the sequential prediction of the transcription and translation sequences, which is more in line with human cognitive principles. By pre-training the decoder, we can directly make better use of the parallel data of MT. Additionally, CTC auxiliary loss, and \textit{shrinking} operation strategies are adopted to enhance our method benefiting from the flexible structure. Experimental results prove the effectiveness of our framework and it has great prospects for promoting the application of speech translation.

\section{Acknowledgements}
\label{sec:acknow}

We'd like to express our gratitude to the anonymous reviewers for their insightful comments. We'd also want to express our gratitude to Rong Ye, Chengqi Zhao, Cheng Yi, and Chi Han for their helpful suggestions and assistance with our project. 
Our research was funded by the Chinese Academy of Sciences' Key Programs under Grant No.ZDBS-SSW-JSC006-2.

\bibliography{aaai21}

\end{document}